\documentclass[10pt,twocolumn,letterpaper]{article}

\usepackage[pagenumbers]{wacv} 

\usepackage{graphicx}
\usepackage{graphics}
\usepackage{amsmath}
\usepackage{amssymb}
\usepackage{booktabs}
\usepackage{bbm}
\usepackage{mathtools}
\usepackage{arydshln}
\usepackage{multirow}
\usepackage{color,soul}
\usepackage{tikz}

\usepackage[
breaklinks,
]{hyperref}

\usepackage[capitalize]{cleveref}
\crefname{section}{Sec.}{Secs.}
\crefname{section}{Section}{Sections}
\crefname{table}{Table}{Tables}
\crefname{table}{Tab.}{Tabs.}

\def\wacvPaperID{909} 
\def\confName{WACV}
\def\confYear{2024}

\begin{document}

\title{Evidential Uncertainty Quantification: A Variance-Based Perspective}

\author{
    Ruxiao Duan$^1$,
    Brian Caffo$^1$,
    Harrison X. Bai$^2$,
    Haris I. Sair$^2$,
    Craig Jones$^1$
    \\
    $^1$Johns Hopkins University
    \\
    $^2$Johns Hopkins University School of Medicine
    \\
    {\tt\small \{rduan6, bcaffo1, hbai7, craigj\}@jhu.edu,
    hsair1@jhmi.edu
    }
}
\maketitle

\begin{abstract}
   Uncertainty quantification of deep neural networks has become an active field of research and plays a crucial role in various downstream tasks such as active learning. Recent advances in evidential deep learning shed light on the direct quantification of aleatoric and epistemic uncertainties with a single forward pass of the model. Most traditional approaches adopt an entropy-based method to derive evidential uncertainty in classification, quantifying uncertainty at the sample level. However, the variance-based method that has been widely applied in regression problems is seldom used in the classification setting. In this work, we adapt the variance-based approach from regression to classification, quantifying classification uncertainty at the class level. The variance decomposition technique in regression is extended to class covariance decomposition in classification based on the law of total covariance, and the class correlation is also derived from the covariance. Experiments on cross-domain datasets are conducted to illustrate that the variance-based approach not only results in similar accuracy as the entropy-based one in active domain adaptation but also brings information about class-wise uncertainties as well as between-class correlations. The code is available at https://github.com/KerryDRX/EvidentialADA. This alternative means of evidential uncertainty quantification will give researchers more options when class uncertainties and correlations are important in their applications.
\end{abstract}


\section{Introduction}

Recent decades have witnessed significant advances in deep learning, which has been employed in various domains including computer vision and natural language processing. Despite the great success of Deep Neural Networks (DNNs), their application in safety-critical tasks is still highly restricted because of their lack of transparency \cite{2022UncertaintySurvey2}, vulnerability to domain shifts \cite{2022UncertaintySurvey13} and adversarial attacks \cite{2022UncertaintySurvey17,2022UncertaintySurvey18,2022UncertaintySurvey19}, and incapability of reliably calibrating prediction uncertainties \cite{2022UncertaintySurvey14}. Therefore, uncertainty quantification of DNNs has attracted great attention more recently, and many approaches \cite{2022UncertaintySurvey30,2022UncertaintySurvey20,2022UncertaintySurvey31,PUE,RENNs,2022UncertaintySurvey34,2022UncertaintySurvey35,2022UncertaintySurvey36} have been devised to estimate predictive uncertainties.

Evidential Deep Learning (EDL) \cite{EDL} brings insights into the direct estimation of model prediction uncertainties with a single forward propagation, and fundamental theories have been established for both classification \cite{EDL,PUE} and regression \cite{DER}. Learning is formulated as a process of evidence acquisition, and each training sample adds support to the higher-order evidential distribution \cite{DER}. Compared to many other alternatives of uncertainty quantification methods, EDL does not modify model architecture or incur additional computation \cite{2022UncertaintySurvey}. In addition, EDL is also able to disentangle uncertainties from different origins: uncertainty due to the conflict of evidence and uncertainty due to the lack of evidence. The former is known as data uncertainty or aleatoric uncertainty (AU), which refers to the intrinsic randomness of the data as a result of natural data complexity and cannot be decreased by training with more data. The latter is known as model uncertainty or epistemic uncertainty (EU), which originates from the lack of knowledge about the data mainly due to data distribution mismatch and can be reduced by increasing training data.

The approach to separate total uncertainty into AU and EU varies in regression and classification. In regression problems \cite{DER,StereoMatching,Molecular}, the total uncertainty is quantified as predictive variance, which can be decomposed based on the law of total variance \cite{Probability}. In classification problems \cite{PUE,OD,DUC}, the total uncertainty is quantified as predictive entropy, which can be decomposed based on the definition of mutual information \cite{InformationTheory}. These two types of methods are respectively referred to as variance-based approach and entropy-based approach in this paper.

Although the entropy-based approach for evidential uncertainty quantification has demonstrated promising results \cite{OD,DUC,BrainTumor}, the variance-based method is typically not considered in classification settings.
Since variance is quantified for a single variable only and is not directly applicable to a probability distribution, entropy has been regarded as the better option to quantify the total uncertainty of a single classification sample. Consequently, classification uncertainty quantification has been limited to the sample level.
However, the variance-based uncertainty quantification introduced in this paper can be applied to each class individually, estimating uncertainty at the class level.

Active domain adaptation (ADA), the combination of active learning (AL) and domain adaptation (DA), is selected as the objective task to evaluate the variance-based uncertainty quantification approach in this work.
DNNs are known to suffer from insufficient annotated data and domain shifts, thus AL and DA have been intensively studied in recent years to alleviate these two problems. 
AL identifies valuable data from the unlabeled pool to annotate, giving the model the best performance with the fewest annotations \cite{SurveyDAL}, but a sampling heuristic for data selection is required, where uncertainty quantification can play an important role. DA mitigates performance degradation when the model is trained on data from one distribution (source domain) but evaluated on data from another distribution (target domain), and a recent study has shown the effectiveness of uncertainty-guided training in unsupervised DA \cite{DUC}. Therefore, the quality of uncertainty quantification can be directly reflected by the performance of ADA.
Dirichlet-based Uncertainty Calibration (DUC) \cite{DUC}, an EDL-based ADA framework, has been proposed and achieved state-of-the-art (SOTA) performance on multiple ADA datasets.

\textbf{Contributions:}
1) While the previous entropy-based approach quantifies EDL uncertainties at the sample level only, we derive a variance-based approach to achieve sample-level and class-level uncertainty quantification.
2) We introduce an EDL class covariance calculation method, based on which between-class correlations can be obtained and highly similar class pairs can be identified.
3) We devise a simultaneous certainty and uncertainty sampling strategy in active learning to boost the model performance under large domain shift.
4) Experiments on cross-domain image classification datasets are conducted to illustrate the effectiveness of our variance-based approach in ADA, with SOTA performance. Class-level uncertainty quantification and class correlation analysis are also demonstrated.

\section{Related Work}

\textbf{Evidential deep learning} \cite{PUE,EDL,DER} treats model learning as an evidence-acquisition process, in which each training sample introduces new evidence to a learned higher-order evidential distribution. The evidential model predicts the parameters of the evidential distribution, then aleatoric and epistemic uncertainties can be directly derived from the distribution parameters. EDL has been applied in various fields such as active learning \cite{DEAL}, domain adaptation \cite{DUC}, semi-supervised learning \cite{SSL}, open set action recognition \cite{OSAR}, reinforcement learning \cite{RL}, object detection \cite{OD}, depth estimation \cite{DER}, stereo matching \cite{StereoMatching}, molecular property prediction \cite{Molecular}, and medical image segmentation \cite{BrainTumor}.

As for uncertainty quantification in EDL, typically the entropy-based approach is adopted in classification \cite{PUE,DUC,SSL,OD,DUC}, while the variance-based approach is used in regression \cite{DER,StereoMatching,Molecular}.
A recent work \cite{DQU} explores the possibility of applying the variance-based approach in a brain segmentation problem, but the quality of uncertainty is not being tested, thus the reliability of this approach in downstream tasks, such as active learning which requires precise uncertainty estimation, remains unclear.

\textbf{Active domain adaptation} is a particular AL problem in which labeled and unlabeled data are drawn from different underlying distributions, thus the representativeness of the target domain (\ie, targetness) is a crucial factor that has to be considered compared to general active learning settings. AADA \cite{13-su2020active} and TQS \cite{14-fu2021transferable} employ a domain discriminator to quantify targetness, but the learning of the discriminator is not related to the classifier, leading to the selection of less useful samples. CLUE \cite{12-prabhu2021active} and DBAL \cite{15-de2021discrepancy} measure targetness by clustering, but redundant sampling becomes a concern as the interrelationship among unlabeled target data is not taken into account.
The domain discriminator and clustering applied in these approaches also lead to extra computational expenses. EADA \cite{16-xie2022active} and SDM-AG \cite{17-xie2022learning} attempt to mitigate domain shift by picking samples based on an auxiliary loss function, but uncertainty quantification is still based on point estimate predictions, which are error-prone on the target dataset. DUC \cite{DUC} leverages evidential uncertainties for model training and sample selection, achieving SOTA performance on multiple ADA datasets.

\section{Evidential Uncertainty Quantification}

\subsection{Previous Work}

\textbf{Deep learning classification.} In a traditional $C$-class classification problem, let $\boldsymbol{y}$ denote the one-hot vector encoding the ground truth class $c_g \in \{1, 2, \dots, C\}$ of input data $\boldsymbol{x}$, with $y_{c_g} = 1$ and $y_c = 0$ for all $c \neq c_g$. A DNN uses softmax activation to convert the prediction logits into a class probability vector $\boldsymbol{\mu} = [\mu_1, \mu_2, \dots, \mu_C]^\top$, in which $\mu_c$ is the probability that input $\boldsymbol{x}$ belongs to class $c$, $\mu_c \in [0, 1]$ for all $c$, and $\sum_{c=1}^{C}\mu_c=1$. Hence,
\begin{equation}
\boldsymbol{y} \sim \mathrm{Cat}(\boldsymbol{\mu})
\label{eq:cls_level1dist}
\end{equation}
where $\mathrm{Cat}(\cdot)$ represents the categorical distribution.

\textbf{Evidential deep learning classification.} EDL \cite{EDL,PUE} considers the class probability vector $\boldsymbol{\mu}$ as a random vector that follows a Dirichlet ($\mathrm{Dir}$) distribution, \ie,
\begin{equation}
\boldsymbol{\mu} \sim \mathrm{Dir}(\boldsymbol{\alpha})
\label{eq:cls_level2dist}
\end{equation}
in which $\boldsymbol{\alpha} = [\alpha_1, \alpha_2, \dots, \alpha_C]^\top$ is a vector of Dirichlet parameters with $\alpha_c > 0$ for all $c$. The probability density function of Dirichlet distribution is given by
\begin{equation}
\mathrm{D}(\boldsymbol{\mu}|\boldsymbol{\alpha}) =  \frac{\Gamma(\alpha_0)}{\prod_{c=1}^{C} \Gamma(\alpha_c)} \prod_{c=1}^C \mu_c^{\alpha_c-1}
\end{equation}
in which $\Gamma(\cdot)$ denotes \emph{Gamma} function and $\alpha_0 = \sum_{c=1}^{C}\alpha_c$ is the Dirichlet strength. Following the Dirichlet distribution, the expected probability vector is $\bar{\boldsymbol{\mu}} := \mathbb{E}[\boldsymbol{\mu}] = \frac{\boldsymbol{\alpha}}{\alpha_0}$, and the expected probability of class $c$ is $\bar{\mu}_c := \mathbb{E}[\mu_c] = \frac{\alpha_c}{\alpha_0}$.

\textbf{Evidential neural network.} Unlike a DNN which predicts class probabilities $\boldsymbol{\mu}$, an evidential neural network (ENN) predicts Dirichlet parameters $\boldsymbol{\alpha}$ instead. Therefore, to construct an ENN, it suffices to replace the softmax activation of a DNN with some function that outputs positive values only (\eg, exponential function) to accommodate the constraint of $\boldsymbol{\alpha}$, and the internal model architecture can be preserved. The final predicted class $c_p$ of an ENN is the class with maximum expected probability, \ie,
\begin{equation}
    c_p = \arg\max_{c} \bar{\mu}_c = \arg\max_{c} \alpha_c
\label{eq:pred}
\end{equation}

\textbf{Evidential model training.} An ENN can be optimized via the loss function $\mathcal{L}^{EDL}$, which consists of a negative log likelihood term $\mathcal{L}^{NLL}$ and a KL-divergence term $\mathcal{L}^{KL}$ for regularization, with $\lambda_{reg}$ being the regularization coefficient.
\begin{equation}
\mathcal{L}^{EDL} = \mathcal{L}^{NLL} + \lambda_{reg} \mathcal{L}^{KL}
\label{eq:edl}
\end{equation}
The final expressions of $\mathcal{L}^{NLL}$ and $\mathcal{L}^{KL}$ are respectively
\begin{flalign}
\mathcal{L}^{NLL} &= \sum_{c=1}^{C} y_c (\log(\alpha_0) - \log(\alpha_c))
\\
\mathcal{L}^{KL} &=
\log \left(
\frac{
    \Gamma (\sum_{c=1}^C \tilde{\alpha}_c )
}{
    \Gamma(C) \prod_{c=1}^C
    \Gamma( \tilde{\alpha}_c )
}
\right)
\nonumber \\
&+
\sum_{c=1}^C (\tilde{\alpha}_c - 1)
\left(
\psi(\tilde{\alpha}_c)
- \psi \left( \sum_{c=1}^C \tilde{\alpha}_c \right)
\right)
\end{flalign}
in which $\tilde{\alpha}_c = y_c + (1 - y_c) \alpha_c$ and $\psi(\cdot)$ is \emph{digamma} function. The derivations are detailed in \cite{EDL}.

\textbf{Entropy-based classification uncertainty.} Previous approaches \cite{OD,DUC,BrainTumor} adopt an entropy-based method to estimate classification uncertainty in EDL, quantifying the total uncertainty $U$ of a sample $\boldsymbol{x}$ as the Shannon entropy of the expected class probabilities of $\boldsymbol{x}$. 
An information theory approach is employed to decompose $U$ into aleatoric uncertainty $U^{alea}$ and epistemic uncertainty $U^{epis}$, \ie, $U = U^{alea} + U^{epis}$. The final expressions of the entropy-based uncertainties are
\begin{flalign}
U &= -\sum_{c=1}^{C} \bar{\mu}_c \log\bar{\mu}_c
\label{eq:ent_U}
\\
U^{alea} &= \sum_{c=1}^{C} \bar{\mu}_c (\psi(\alpha_0+1) - \psi(\alpha_c+1))
\label{eq:ent_Ualea}
\\
U^{epis}
    &= -\sum_{c=1}^{C} \bar{\mu}_c (\log\bar{\mu}_c + \psi(\alpha_0+1) - \psi(\alpha_c+1))
\label{eq:ent_Uepis}
\end{flalign}
The derivations can be found in \cite{PUE}.

\textbf{Variance-based regression uncertainty.} Deep evidential regression \cite{DER} has been proposed for regression uncertainty quantification. Formally, it is assumed that the prediction target $y$ is a continuous random variable following a Gaussian distribution, whose mean and variance jointly follow a Normal Inverse-Gamma (NIG) distribution:
\begin{flalign}
y &\sim \mathcal{N}(\mu, \sigma^2)
\label{eq:reg_level1dist}
\\
\mu, \sigma^2 &\sim \mathrm{NIG}(\gamma, \nu, \alpha, \beta)
\label{eq:reg_level2dist}
\end{flalign}
in which $\gamma \in \mathbb{R}, \nu>0, \alpha>1, \beta>0$. Thus a regression ENN predicts the NIG distribution parameters.

In regression, the total uncertainty $U$ is quantified as the variance of prediction target $y$. Based on the law of total variance \cite{Probability}, the total uncertainty can be decomposed into aleatoric and epistemic components:
\begin{equation}
\label{eq:var_decompose}
    \underbrace{\mathrm{Var}[y]}_\text{total}
    = \underbrace{\mathbb{E}[\mathrm{Var}[y | \mu, \sigma^2]]}_\text{aleatoric}
    + \underbrace{\mathrm{Var}[\mathbb{E}[y | \mu, \sigma^2]]}_\text{epistemic}
\end{equation}
According to \cite{DER}, the regression uncertainties are derived as
\begin{flalign}
U &= \frac{(\nu+1)\beta}{\nu(\alpha-1)}
\\
U^{alea} &= \frac{\beta}{\alpha-1} \label{eq:reg_alea}
\\
U^{epis} &= \frac{\beta}{\nu(\alpha-1)} \label{eq:reg_epis}
\end{flalign}

\subsection{Variance-based Classification Uncertainty}

\textbf{Motivation.} A weakness of the traditional entropy-based approach is that classification uncertainties $U$, $U^{alea}$, and $U^{epis}$ can only be quantified up to the sample level (\cref{eq:ent_U}, \cref{eq:ent_Ualea}, \cref{eq:ent_Uepis}). That is to say, we can at best derive uncertainties associated with an entire single sample $\boldsymbol{x}$. Imagine a scenario in which the predictive uncertainty of a vehicle image $\boldsymbol{x}$ is high because the model is not able to distinguish between two classes on this image: automobile and truck. For most other categories, such as cat and dog, the model is certain that the image does not belong to any of them, thus those classes do not contribute much to the total uncertainty. The entropy-based approach does not provide any information about these class-level uncertainties, which can sometimes be crucial in downstream applications.

\textbf{From regression to classification.} In EDL, classification and regression problems share the same underlying assumption: the prediction target is supposed to follow a bi-level probability distribution (\cref{eq:cls_level1dist} and \cref{eq:cls_level2dist} for classification; \cref{eq:reg_level1dist} and \cref{eq:reg_level2dist} for regression). Therefore, the uncertainty quantification and decomposition approach in regression (\cref{eq:var_decompose}) should also be applicable in classification. However, regression has a scalar target $y$ while classification has a vector target $\boldsymbol{y}$. Since variance is not directly calculable for a multidimensional vector, we extend the variance in \cref{eq:var_decompose} to covariance to adapt to the multidimensionality of $\boldsymbol{y}$.

\textbf{Covariance quantification and decomposition.}
The covariance matrix of a classification target label $\boldsymbol{y}$ is defined as
\begin{equation}
    \mathrm{Cov}[\boldsymbol{y}] := \mathbb{E}[(\boldsymbol{y}-\mathbb{E}[\boldsymbol{y}])(\boldsymbol{y}-\mathbb{E}[\boldsymbol{y}])^\top]
\end{equation}
which can be decomposed into aleatoric and epistemic components based on the law of total covariance \cite{Probability}:
\begin{equation}
\label{eq:cov_decompose}
    \underbrace{\mathrm{Cov}[\boldsymbol{y}]}_\text{total}
    = \underbrace{\mathbb{E}[\mathrm{Cov}[\boldsymbol{y} | \boldsymbol{\mu}]]}_\text{aleatoric}
    + \underbrace{\mathrm{Cov}[\mathbb{E}[\boldsymbol{y} | \boldsymbol{\mu}]]}_\text{epistemic}
\end{equation}

We define the aleatoric and epistemic components of the covariance matrix $\mathrm{Cov}[\boldsymbol{y}]$ as aleatoric covariance matrix $\mathrm{Cov}[\boldsymbol{y}]^{alea}$ and epistemic covariance matrix $\mathrm{Cov}[\boldsymbol{y}]^{epis}$. Following the distribution assumptions \cref{eq:cls_level1dist} and \cref{eq:cls_level2dist}, the covariance matrices can be derived as follows:
\begin{flalign}
\mathrm{Cov}[\boldsymbol{y}]
    &= \mathrm{Diag}(\bar{\boldsymbol{\mu}}) - \bar{\boldsymbol{\mu}} \bar{\boldsymbol{\mu}}^\top \label{eq:Cov}
    \\
\mathrm{Cov}[\boldsymbol{y}]^{alea}
    &= \frac{\alpha_0}{\alpha_0+1}
    (
    \mathrm{Diag}(\bar{\boldsymbol{\mu}}) - \bar{\boldsymbol{\mu}} \bar{\boldsymbol{\mu}}^\top
    ) \label{eq:Cov_alea}
    \\
\mathrm{Cov}[\boldsymbol{y}]^{epis}
    &= \frac{1}{\alpha_0+1}
    (
    \mathrm{Diag}(\bar{\boldsymbol{\mu}}) - \bar{\boldsymbol{\mu}} \bar{\boldsymbol{\mu}}^\top
    ) \label{eq:Cov_epis}
\end{flalign}
in which $\mathrm{Diag}(\cdot)$ represents a diagonal matrix with the specified vector on its diagonal. The derivations are provided in our supplementary material.

Similar to regression (\cref{eq:reg_alea} and \cref{eq:reg_epis}), the aleatoric and epistemic components are proportional: $\frac{\mathrm{Cov}[\boldsymbol{y}]^{alea}}{\mathrm{Cov}[\boldsymbol{y}]^{epis}} = \alpha_0$, which is reasonable since as the model gains more evidence, the Dirichlet strength $\alpha_0$ tends to increase, suppressing the model uncertainty.

\textbf{Class-wise uncertainty quantification.} With the covariance matrices, the variance-based evidential uncertainties of each class can be directly retrieved from the diagonal:
\begin{flalign}
U_c
&= \mathrm{Cov}[\boldsymbol{y}]_{c,c} 
= \bar{\mu}_c (1 - \bar{\mu}_c)
\label{eq:var_classU}
\\
U_c^{alea}
&= \mathrm{Cov}[\boldsymbol{y}]_{c,c}^{alea}
= \frac{\alpha_0}{\alpha_0+1} \bar{\mu}_c (1 - \bar{\mu}_c)
\label{eq:var_classUalea}
\\
U_c^{epis}
&= \mathrm{Cov}[\boldsymbol{y}]_{c,c}^{epis}
= \frac{1}{\alpha_0+1} \bar{\mu}_c (1 - \bar{\mu}_c)
\label{eq:var_classUepis}
\end{flalign}
in which $U_c$, $U_c^{alea}$, and $U_c^{epis}$ are the total, aleatoric, and epistemic uncertainties associated with class $c$ of data $\boldsymbol{x}$.

\textbf{Sample-wise uncertainty quantification.} Sample-wise uncertainty can be obtained by simply aggregating the class-wise uncertainties over all the classes:
\begin{flalign}
U
&= \sum_{c=1}^{C} U_c
= 1 - \sum_{c=1}^{C} \bar{\mu}_c^2
\label{eq:var_sampleU}
\\
U^{alea}
&= \sum_{c=1}^{C} U_c^{alea}
= \frac{\alpha_0}{\alpha_0+1} \left(1 - \sum_{c=1}^{C} \bar{\mu}_c^2\right)
\label{eq:var_sampleUalea}
\\
U^{epis}
&= \sum_{c=1}^{C} U_c^{epis}
= \frac{1}{\alpha_0+1} \left(1 - \sum_{c=1}^{C} \bar{\mu}_c^2\right)
\label{eq:var_sampleUepis}
\end{flalign}
in which $U$, $U^{alea}$, and $U^{epis}$ are the total, aleatoric, and epistemic uncertainties associated with data $\boldsymbol{x}$.

\textbf{Correlation quantification.} The class correlation matrix can be directly obtained from the covariance matrix by definition:
\begin{equation}
\mathrm{Corr}[\boldsymbol{y}] := \frac{\mathrm{Cov}[\boldsymbol{y}]}{{{\sigma}(\boldsymbol{y})}{{{\sigma}(\boldsymbol{y})}}^\top}
\label{eq:corr}
\end{equation}
in which ${\sigma}(\boldsymbol{y}) = \sqrt{\mathrm{diag}(\mathrm{Cov}[\boldsymbol{y}])}$ represents the class standard deviation and $\mathrm{diag}(\cdot)$ denotes the diagonal of a matrix. The off-diagonal entries of the correlation matrix provide between-class correlation information, \eg, $\mathrm{Corr}[\boldsymbol{y}]_{i,j}$ is the correlation between class $i$ and class $j$ ($i \neq j$) on the sample $\boldsymbol{x}$. This gives a measure of how different classes correlate with each other in a given sample.

\textbf{Comparison with entropy-based approach.} Entropy-based and variance-based approaches adopt the same way to construct and train the ENN but quantify uncertainties in different manners. Both methods can quantify uncertainties of a sample $\boldsymbol{x}$ as a whole (\cref{eq:ent_U}, \cref{eq:ent_Ualea}, \cref{eq:ent_Uepis} for entropy-based; \cref{eq:var_sampleU}, \cref{eq:var_sampleUalea}, \cref{eq:var_sampleUepis} for variance-based). However, our variance-based approach further provides uncertainties of each single class $c$ of sample $\boldsymbol{x}$ (\cref{eq:var_classU}, \cref{eq:var_classUalea}, \cref{eq:var_classUepis}) and yields the between-class correlation of each pair of classes (\cref{eq:corr}).

\textbf{From theory to application.}
The variance-based classification EDL theories can be applied following three steps:
1) Choose an appropriate classification DNN model and turn it into an ENN by changing the output activation from softmax to any function that produces only positive values (\eg, exponential function).
2) Train the ENN by \cref{eq:edl}.
3) Given a test sample $\boldsymbol{x}$ during inference, the ENN predicts the Dirichlet parameters $\boldsymbol{\alpha}$. Then the class prediction is given by \cref{eq:pred}, the associated predictive uncertainties can be quantified via \cref{eq:var_classU}-\cref{eq:var_sampleUepis}, and the class correlations can be calculated with \cref{eq:corr}.

\section{Active Domain Adaptation}

\subsection{Previous Work}

\textbf{Problem formulation.} ADA assumes that there is a source domain dataset $\mathcal{S}=\{(\boldsymbol{x}_i^s, \boldsymbol{y}_i^s)\}_{i=1}^{n_s}$ and a target domain dataset $\mathcal{T}=\{\boldsymbol{x}_i^t\}_{i=1}^{n_t}$. The two domains follow different data distributions but share the same label space. Target data are separated into a labeled subset $\mathcal{T}^l$ and an unlabeled subset $\mathcal{T}^u$, \ie, $\mathcal{T} = \mathcal{T}^l \cup \mathcal{T}^u$. Initially, all target data are unlabeled, $\mathcal{T}^l =$ \O \:and $\mathcal{T}^u = \mathcal{T}$.

In active selection round $k \in \{1, 2, \dots, K\}$, $b_k^u$ samples in $\mathcal{T}^u$ are selected by some sampling strategy to query labels from the oracle. These selected data are then removed from $\mathcal{T}^u$ and added to $\mathcal{T}^l$ with their labels. A total budget of $B$ determines the maximum number of samples that can be labeled, and typically, $B=\sum_{k=1}^{K} b_k^u$.

\textbf{DUC algorithm.} DUC \cite{DUC} is an EDL-based active learning framework that has achieved SOTA performance in ADA. Its active sampling approach employs a two-step selection technique based on two types of evidential uncertainties. In each active round $k$, the data from $\mathcal{T}^u$ are first sorted by their EU. The top $\kappa b_k^u$ uncertain samples are passed to the second step and sorted by their AU. The most uncertain $b_k^u$ samples in the second step are finally sent to the oracle to annotate.

\textbf{DUC model training.} In addition to supervised training of EDL, unsupervised uncertainty guidance (UG) \cite{DUC} is utilized in model optimization. By minimizing the uncertainties of unlabeled targets $\mathcal{T}^u$, domain shift can be greatly reduced. The unsupervised loss for a sample $\boldsymbol{x}$ is designed as the combination of its AU and EU:
\begin{equation}
\mathcal{L}^{UG} = \lambda_{a}U^{alea} + \lambda_{e}U^{epis}
\end{equation}
The total loss is the sum of supervised loss (EDL) and unsupervised loss (UG).
\begin{equation}
\mathcal{L} 
= \sum_{\boldsymbol{x}_i \in \mathcal{S} \cup \mathcal{T}^l} \mathcal{L}^{EDL}(\boldsymbol{x}_i)
+ \sum_{\boldsymbol{x}_i \in \mathcal{T}^u} \mathcal{L}^{UG}(\boldsymbol{x}_i)
\end{equation}

\subsection{Certainty Sampling}

\textbf{Motivation.} In this paper, we propose simultaneous certainty and uncertainty sampling in active learning to exploit the value of confident samples. Previous AL and ADA algorithms mainly focus on picking the most uncertain samples to annotate, ignoring the value of certain samples in the dataset. However, if wisely utilized, certain samples can boost the model performance:
1) As the model is certain about those data, their predictions are very likely to be correct. Thus adding them to the labeled pool can provide additional information for supervised learning.
2) Removing certain samples from the unlabeled pool effectively reduces the search space for uncertain samples. Thus for uncertainty sampling in the current and subsequent active rounds, computation is reduced and the likelihood of discovering informative samples is increased.

\begin{figure}[t]
  \centering
   \includegraphics[width=1\linewidth]{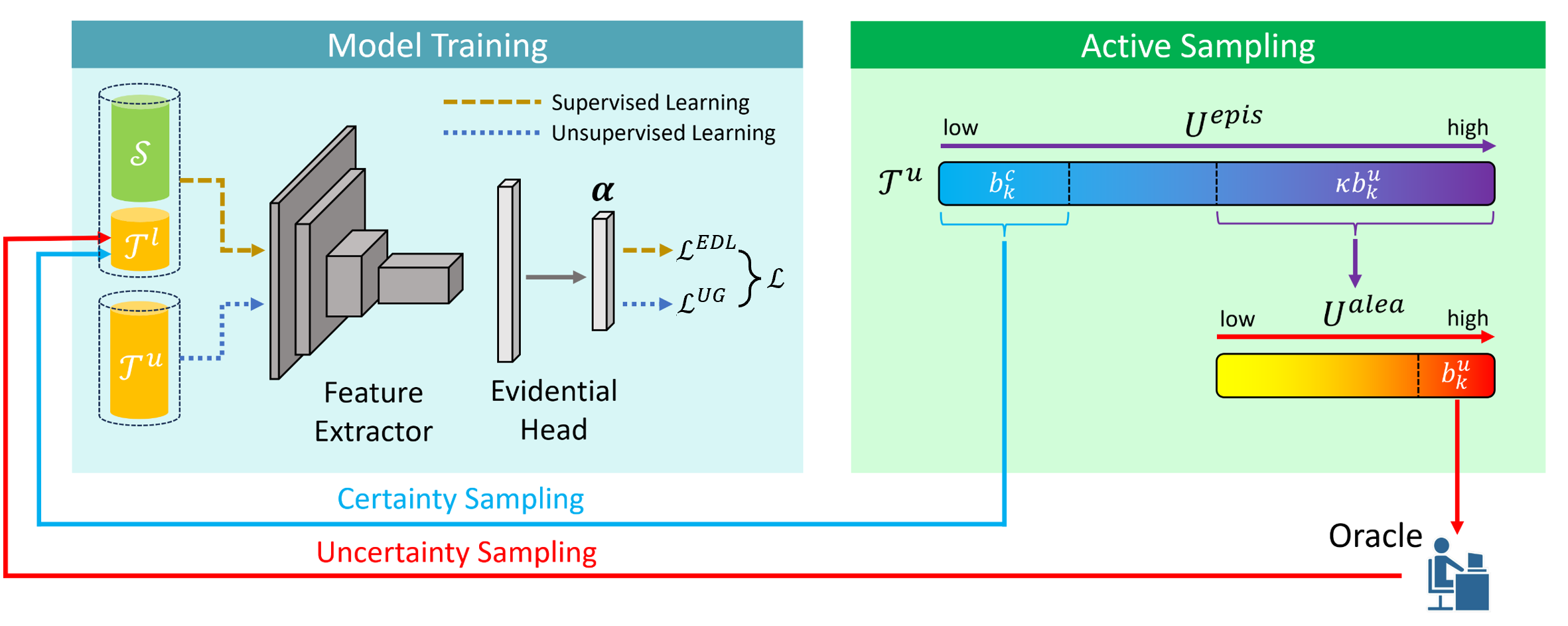}
   \caption{DUC framework with simultaneous certainty and uncertainty sampling. In active sampling round $k$, the unlabeled target data in $\mathcal{T}^u$ are first sorted by their epistemic uncertainty $U^{epis}$. The $b_k^c$ samples with the least $U^{epis}$ are moved from $\mathcal{T}^u$ to labeled target dataset $\mathcal{T}^l$ with their pseudo labels, and the $\kappa b_k^u$ samples with the most $U^{epis}$ are then sorted by their aleatoric uncertainty $U^{alea}$. The $b_k^u$ samples with the most $U^{alea}$ are sent to the oracle to query labels, after which these samples are moved to $\mathcal{T}^l$.}
   \label{fig:DUC}
\end{figure}

\textbf{DUC with certainty sampling.} As illustrated in \cref{fig:DUC}, certainty sampling can be incorporated into DUC in its first step of active selection: when sorting $\mathcal{T}^u$ by their EU, we select the top $b_k^c$ samples with the least EU as the certain samples and add them directly to $\mathcal{T}^l$ with their predictions as pseudo labels.
Note that this EU-based certainty sampling does not incur extra computation, as $\mathcal{T}^u$ is not sorted one more time using another metric.
Though applied here in the DUC framework, this idea of certainty sampling is generally applicable in any AL algorithm.

\textbf{Class-balanced certainty sampling.} We introduce a variant of certainty sampling to address the concern that the selected certain samples might be biased towards simple classes. To avoid training set imbalance, when picking $b_k^c$ most certain samples in round $k$, we first divide $\mathcal{T}^u$ into $C$ classes according to their predictions and then select about $\frac{b_k^c}{C}$ most certain samples from each class. The class-balanced selection is vital if class difficulties vary greatly.

\section{Experiments}

\subsection{Datasets}

Experiments are based on two multi-domain image classification datasets: Office-Home \cite{OfficeHome} and Visda-2017 \cite{Visda2017}.

Office-Home \cite{OfficeHome} is a set of 15,588 images categorized into 65 classes that are collected in typical office and home settings. Images are drawn from 4 different domains: Art (A), Clipart (C), Product (P), and Real World (R).

Visda-2017 \cite{Visda2017} contains synthetic (S) and real (R) images categorized into 12 classes. We use the training set (152,397 synthetic images) as source data and the validation set (55,388 real images) as target data.

\begin{table*}[t]
\centering
\resizebox{\textwidth}{!}{%
\begin{tabular}{c | c | c c c c c c c c c c c c | c}
    \hline \hline
    \multirow{2}{*}{AL Method}
    & Visda-2017
    & \multicolumn{13}{c}{Office-Home}
    \\ \cline{2-15}
    {}
    & S$\rightarrow$R
    & A$\rightarrow$C
    & A$\rightarrow$P
    & A$\rightarrow$R
    & C$\rightarrow$A
    & C$\rightarrow$P
    & C$\rightarrow$R
    & P$\rightarrow$A
    & P$\rightarrow$C
    & P$\rightarrow$R
    & R$\rightarrow$A
    & R$\rightarrow$C
    & R$\rightarrow$P
    & Average \\
    \hline
    Source-only
    &44.7$\pm$0.1 &42.1 &66.3 &73.3 &50.7 &59.0 &62.6 &51.9 &37.9 &71.2 &65.2 &42.6 &76.6 &58.3
    \\ \hline
    CoreSet \cite{11-sener2017active}
    &81.9$\pm$0.3 &51.8 &72.6 &75.9 &58.3 &68.5 &70.1 &58.8 &48.8 &75.2 &69.0 &52.7 &80.0 &65.1
    \\ 
    Random
    &78.1$\pm$0.6 &52.5 &74.3 &77.4 &56.3 &69.7 &68.9 &57.7 &50.9 &75.8 &70.0 &54.6 &81.3 &65.8
    \\
    BvSB \cite{22-joshi2009multi}
    &81.3$\pm$0.4 &56.3 &78.6 &79.3 &58.1 &74.0 &70.9 &59.5 &52.6 &77.2 &71.2 &56.4 &84.5 &68.2
    \\
    WAAL \cite{24-shui2020deep}
    &83.9$\pm$0.4 &55.7 &77.1 &79.3 &61.1 &74.7 &72.6 &60.1 &52.1 &78.1 &70.1 &56.6 &82.5 &68.3
    \\
    Entropy
    &82.7$\pm$0.3 &58.0 &78.4 &79.1 &60.5 &73.0 &72.6 &60.4 &54.2 &77.9 &71.3 &58.0 &83.6 &68.9
    \\
    BADGE \cite{25-ash2019deep}
    &84.3$\pm$0.3 &58.2 &79.7 &79.9 &61.5 &74.6 &72.9 &61.5 &56.0 &78.3 &71.4 &60.9 &84.2 &69.9
    \\ \hline
    AADA \cite{13-su2020active}
    &80.8$\pm$0.4 &56.6 &78.1 &79.0 &58.5 &73.7 &71.0 &60.1 &53.1 &77.0 &70.6 &57.0 &84.5 &68.3
    \\
    DBAL \cite{15-de2021discrepancy}
    &82.6$\pm$0.3 &58.7 &77.3 &79.2 &61.7 &73.8 &73.3 &62.6 &54.5 &78.1 &72.4 &59.9 &84.3 &69.6
    \\
    TQS \cite{14-fu2021transferable}
    &83.1$\pm$0.4 &58.6 &81.1 &81.5 &61.1 &76.1 &73.3 &61.2 &54.7 &79.7 &73.4 &58.9 &86.1 &70.5
    \\
    CLUE \cite{12-prabhu2021active}
    &85.2$\pm$0.4 &58.0 &79.3 &80.9 &68.8 &77.5 &76.7 &66.3 &57.9 &81.4 &75.6 &60.8 &86.3 &72.5
    \\
    EADA \cite{16-xie2022active}
    &88.3$\pm$0.1 &63.6 &84.4 &83.5 &70.7 &83.7 &80.5 &73.0 &63.5 &85.2 &78.4 &65.4 &88.6 &76.7
    \\ \hdashline
    DUC-ENT w/o CS \cite{DUC}
    &89.3$\pm$0.1 &66.2$\pm$0.9 &85.1$\pm$0.3 &84.7$\pm$0.2 &71.1$\pm$1.0 &84.0$\pm$0.4 &81.5$\pm$0.0 &72.0$\pm$0.8 &67.0$\pm$0.8 &85.6$\pm$0.2 &79.9$\pm$0.4 &71.0$\pm$0.5 &89.5$\pm$0.5 &78.1$\pm$0.2
    \\
    DUC-VAR w/o CS
    &89.4$\pm$0.1 &66.2$\pm$0.4 &85.0$\pm$0.4 &85.0$\pm$0.2 &70.8$\pm$0.4 &85.1$\pm$0.3 &81.7$\pm$0.2 &71.7$\pm$0.1 &67.0$\pm$0.3 &85.6$\pm$0.5 &80.1$\pm$0.3 &70.9$\pm$0.3 &89.4$\pm$0.4 &78.2$\pm$0.1
    \\
    DUC-ENT w/ CS
    &89.3$\pm$0.2
    &\textbf{69.0}$\pm$0.7
    &85.3$\pm$0.9
    &\textbf{85.3}$\pm$0.1
    &73.0$\pm$0.3
    &85.1$\pm$0.4
    &82.6$\pm$0.1
    &73.9$\pm$0.3
    &68.5$\pm$0.3
    &\textbf{86.0}$\pm$0.1
    &\textbf{80.6}$\pm$0.5
    &\textbf{72.5}$\pm$0.3
    &89.5$\pm$0.3
    &79.3$\pm$0.2
    \\
    DUC-VAR w/ CS
    &\textbf{89.5}$\pm$0.1
    &68.3$\pm$0.4
    &\textbf{85.9}$\pm$0.2
    &85.2$\pm$0.1
    &\textbf{73.8}$\pm$0.0
    &\textbf{85.6}$\pm$0.4
    &\textbf{83.4}$\pm$0.5
    &\textbf{74.8}$\pm$0.9
    &\textbf{68.9}$\pm$0.5
    &85.7$\pm$0.3
    &80.2$\pm$0.0
    &71.7$\pm$0.9
    &\textbf{89.7}$\pm$0.3
    &\textbf{79.5}$\pm$0.2
    \\ \hline \hline
\end{tabular}
}
\caption{Classification accuracy (\%) of ResNet-50 on Office-Home and Visda-2017 with a budget of 5\% target data. 
DUC-ENT: DUC with entropy-based uncertainty quantification.
DUC-VAR: DUC with variance-based uncertainty quantification.
CS: certainty sampling.
}
\label{tab:1}
\end{table*}

\subsection{Experimental Setup}

We adopt most experimental settings of \cite{DUC} to get comparable results. There are $K = 5$ active selection rounds and a budget of $B = 5\% |\mathcal{T}|$ in total.
In active round $k \in \{1, 2, \dots, K\}$, $\kappa=10$, and $b_k^u = B/5 = 1\% |\mathcal{T}|$ uncertain samples and $b_k^c = k\% |\mathcal{T}|$ certain samples are selected from the unlabeled pool $\mathcal{T}^u$. We gradually increase the certainty sampling ratio because as training progresses, the model can give increasingly more correct predictions.

ENN is constructed using ResNet-50 \cite{ResNet} pre-trained on ImageNet \cite{ImageNet} as the backbone and exponential function as the output activation. The model is trained for 50 epochs on Office-Home and 40 epochs on Visda-2017, and active sampling is performed in epochs 10, 12, 14, 16, and 18.
SGD optimizer with batch size 32, momentum 0.9, and weight decay of 0.001 and the learning rate scheduling policy in \cite{21-long2018conditional} are used for model training.
The initial learning rates for Office-Home and Visda-2017 are 0.004 and 0.001, respectively.
In the loss function, $\lambda_{a} = 0.05$. 
For entropy-based method, $\lambda_{e} = 1$. 
For variance-based method, $\lambda_{e} = 50$ for Office-Home and $\lambda_{e} = 10$ for Visda-2017.
The regularization coefficient $\lambda_{reg} = \frac{1}{C}$ to mitigate the influence of the number of classes $C$.

For each domain transition, we independently run the experiment three times, each with a randomly selected seed, and report the mean and standard deviation of the three final accuracies. The experiments are implemented with PyTorch \cite{PyTorch} on an NVIDIA GeForce RTX 4090 Laptop GPU.

\subsection{Results}

\subsubsection{Classification Accuracy}

The accuracy comparison of DUC with entropy-based and variance-based evidential uncertainties (denoted by DUC-ENT and DUC-VAR respectively) on the two datasets is illustrated in \cref{tab:1}, together with the results of previous AL algorithms.

The variance-based approach achieves slightly better average accuracy than the entropy-based one on the two datasets and is also more stable when certainty sampling is not involved, as the average standard deviation of DUC-VAR on 12 domains of Office-Home (0.32) is significantly lower than that of DUC-ENT (0.50).

With certainty sampling, the average accuracy of Office-Home is improved by more than 1\%, but the effect of certainty sampling on transitions with little domain shift is negligible (\eg, on Visda-2017 S$\rightarrow$R and Office-Home R$\rightarrow$P). However, it is noteworthy that certainty sampling never harms the model performance in any domain transition using either DUC-ENT or DUC-VAR.

\begin{table}[t]
\centering
\resizebox{!}{.05\textheight}{%
\begin{tabular}{c | c | c | c | c}
\hline\hline
    Source & UG & US & CS & A$\rightarrow$C
    \\ \hline
    \checkmark & - & - & - & 46.0
    \\
    \checkmark & \checkmark & - & - & 52.8
    \\
    \checkmark & - & \checkmark & - & 62.1
    \\
    \checkmark & \checkmark & \checkmark & - & 66.2
    \\
    \checkmark & \checkmark & \checkmark & \checkmark & 68.3
    \\ \hline\hline
\end{tabular}
}
\caption{Ablation study on Office-Home A$\rightarrow$C domain transition. Source: training by source data. UG: uncertainty guidance. US: uncertainty sampling. CS: certainty sampling.}
\label{tab:ablation}
\end{table}

\begin{table*}[t]
\centering
\resizebox{\textwidth}{!}{%
\begin{tabular}{c | c | c | c c c c c c c c c c c c | c}
    \hline \hline
    \multirow{2}{*}{Uncertainty}
    & \multirow{2}{*}{Quantification}
    & Visda-2017
    & \multicolumn{13}{c}{Office-Home}
    \\ \cline{3-16}
    {}
    & {}
    & S$\rightarrow$R
    & A$\rightarrow$C
    & A$\rightarrow$P
    & A$\rightarrow$R
    & C$\rightarrow$A
    & C$\rightarrow$P
    & C$\rightarrow$R
    & P$\rightarrow$A
    & P$\rightarrow$C
    & P$\rightarrow$R
    & R$\rightarrow$A
    & R$\rightarrow$C
    & R$\rightarrow$P
    & Average
    \\ \hline
    \multirow{2}{*}{AU} & ENT
    &\textbf{75.6}$\pm$0.2
    &78.9$\pm$0.1
    &80.8$\pm$0.5
    &84.3$\pm$0.1
    &83.4$\pm$0.2
    &83.3$\pm$0.4
    &83.2$\pm$0.7
    &\textbf{83.4}$\pm$0.1
    &80.1$\pm$0.1
    &85.7$\pm$0.1
    &85.7$\pm$0.2
    &\textbf{81.3}$\pm$0.5
    &85.8$\pm$0.0
    &83.0$\pm$0.2
    \\
    {} & VAR
    &75.3$\pm$0.1
    &\textbf{79.5}$\pm$0.2
    &\textbf{81.5}$\pm$0.2
    &\textbf{85.0}$\pm$0.0
    &\textbf{83.5}$\pm$0.4
    &\textbf{83.4}$\pm$0.1
    &\textbf{83.6}$\pm$0.1
    &83.3$\pm$0.2
    &\textbf{80.7}$\pm$0.3
    &\textbf{86.0}$\pm$0.2
    &\textbf{86.3}$\pm$0.2
    &80.0$\pm$0.2
    &\textbf{86.5}$\pm$0.0
    &\textbf{83.3}$\pm$0.1
    \\ \hdashline
    \multirow{2}{*}{EU} & ENT
    &73.8$\pm$0.1
    &77.8$\pm$0.0
    &79.5$\pm$1.0
    &83.1$\pm$0.3
    &81.8$\pm$0.1
    &82.2$\pm$0.4
    &81.8$\pm$0.7
    &81.9$\pm$0.1
    &78.8$\pm$0.0
    &84.4$\pm$0.1
    &83.7$\pm$0.3
    &80.1$\pm$0.6
    &84.5$\pm$0.0
    &81.7$\pm$0.2
    \\
    {} & VAR
    &73.8$\pm$0.0
    &78.5$\pm$0.0
    &80.5$\pm$0.2
    &84.2$\pm$0.1
    &82.3$\pm$0.6
    &81.8$\pm$0.0
    &82.0$\pm$0.2
    &81.8$\pm$0.0
    &79.2$\pm$0.4
    &84.7$\pm$0.1
    &84.9$\pm$0.3
    &78.6$\pm$0.2
    &85.5$\pm$0.1
    &82.0$\pm$0.0
    \\ \hline \hline
\end{tabular}
}
\caption{
AUROC (\%) of misclassification detection on Office-Home and Visda-2017 based on uncertainties in epoch 10.
AU: aleatoric uncertainty.
EU: epistemic uncertainty.
ENT: entropy-based uncertainty quantification.
VAR: variance-based uncertainty quantification.
}
\label{tab:AUC}
\end{table*}

\subsubsection{Ablation Study}

The three main components of the improved DUC framework with variance-based uncertainty quantification are uncertainty guidance (UG), uncertainty sampling (US), and certainty sampling (CS). The contributions of the three parts are analyzed through an ablation study on the Office-Home dataset Art to Clipart domain transition (\cref{tab:ablation}) since this transition suffers from the greatest domain shift and makes the effect of each component more evident.

\textbf{Uncertainty guidance.} It can be noticed that UG training always improves accuracy on the target domain, with or without active US. Even in the unsupervised domain adaptation setting where no target samples are labeled, UG improves accuracy on the target domain by a large margin (6.8\%), meaning that minimizing target evidential uncertainty serves as good guidance to model training.

\begin{figure}[t]
\centering
\includegraphics[scale=0.3]{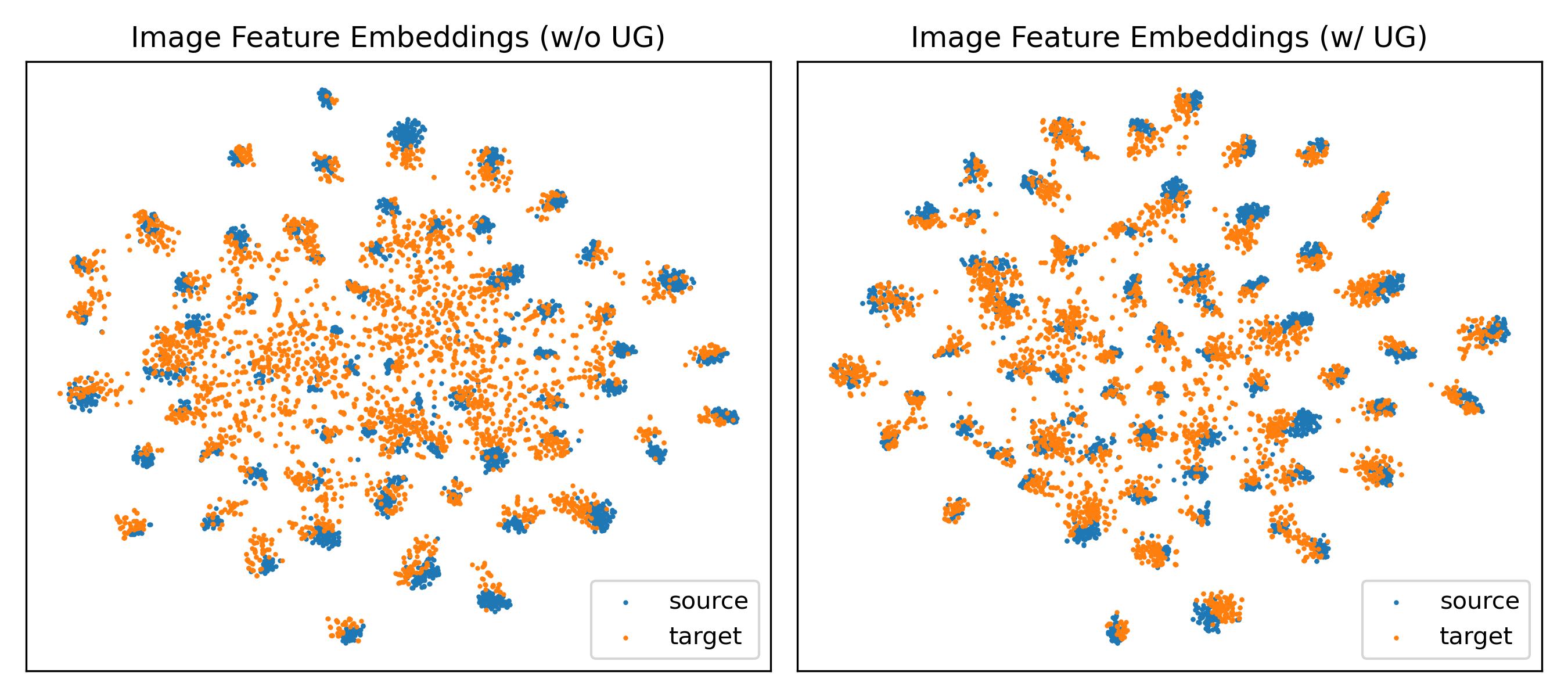}
\caption{
The t-SNE visualization of image feature embeddings derived by the model trained without (left) and with (right) variance-based uncertainty guidance on Office-Home dataset A$\rightarrow$C domain transition. No active sampling is performed.
}
\label{fig:tsne}
\end{figure}

A qualitative study on the feature space (\cref{fig:tsne}) demonstrates that UG training significantly mitigates the domain shift from source to target. When no guidance is provided, target features shift greatly from source ones and tend to be widely spread, incapable of forming into class-wise clusters. However, with the guidance of variance-based evidential uncertainties, the target features appear visually closer to the source ones and learn to group into small clusters. The learned features of target samples from different classes are far more distinguishable, even if no target labels are provided to them at this stage.

\textbf{Certainty sampling.} A vital prerequisite for CS to work is that the certain sample selection criterion should be a strong indicator of prediction correctness. Therefore, whether low epistemic uncertainty indicates a high likelihood of prediction correctness has to be examined. The average prediction accuracy of all the selected class-balanced certain samples is 94.2\% in Office-Home and 97.2\% in Visda-2017. Since a total of 15\% target data are selected as certain samples, prediction mistakes are almost inevitable, but an overall accuracy of more than 94\% is still sufficient to improve the model performance under most circumstances.

\subsubsection{Evidential Uncertainty Analysis}

\textbf{Evidential uncertainty for misclassification detection.} Apparently, it is preferable that before active selection, most samples chosen in US are incorrectly predicted and most samples chosen in CS are correctly predicted, so that we can maximize the performance gain from a limited labeling budget without introducing excessive wrong pseudo labels in training. Both these two desires require a strong relationship between evidential uncertainty and misclassification.

This relationship can be explicitly measured by treating the uncertainty estimation as a binary classification problem, in which the prediction target is whether the sample is misclassified or not and the prediction score is the evidential uncertainty \cite{PUE}. In this way, AUROC can be quantified to evaluate how evidential uncertainty is related to the likelihood of giving a wrong prediction. The mean and standard deviation of AUROC scores from the three runs of DUC are given in \cref{tab:AUC}. The scores are calculated in epoch 10, right before the first active sampling round. It can be observed that variance-based evidential uncertainties have a higher AUROC score than the entropy-based one on most domain transitions in Office-Home, but the entropy-based aleatoric uncertainty is a better metric to detect misclassification than the variance-based one in Visda-2017.

\begin{figure}[t]
\centering
\includegraphics[scale=0.4]{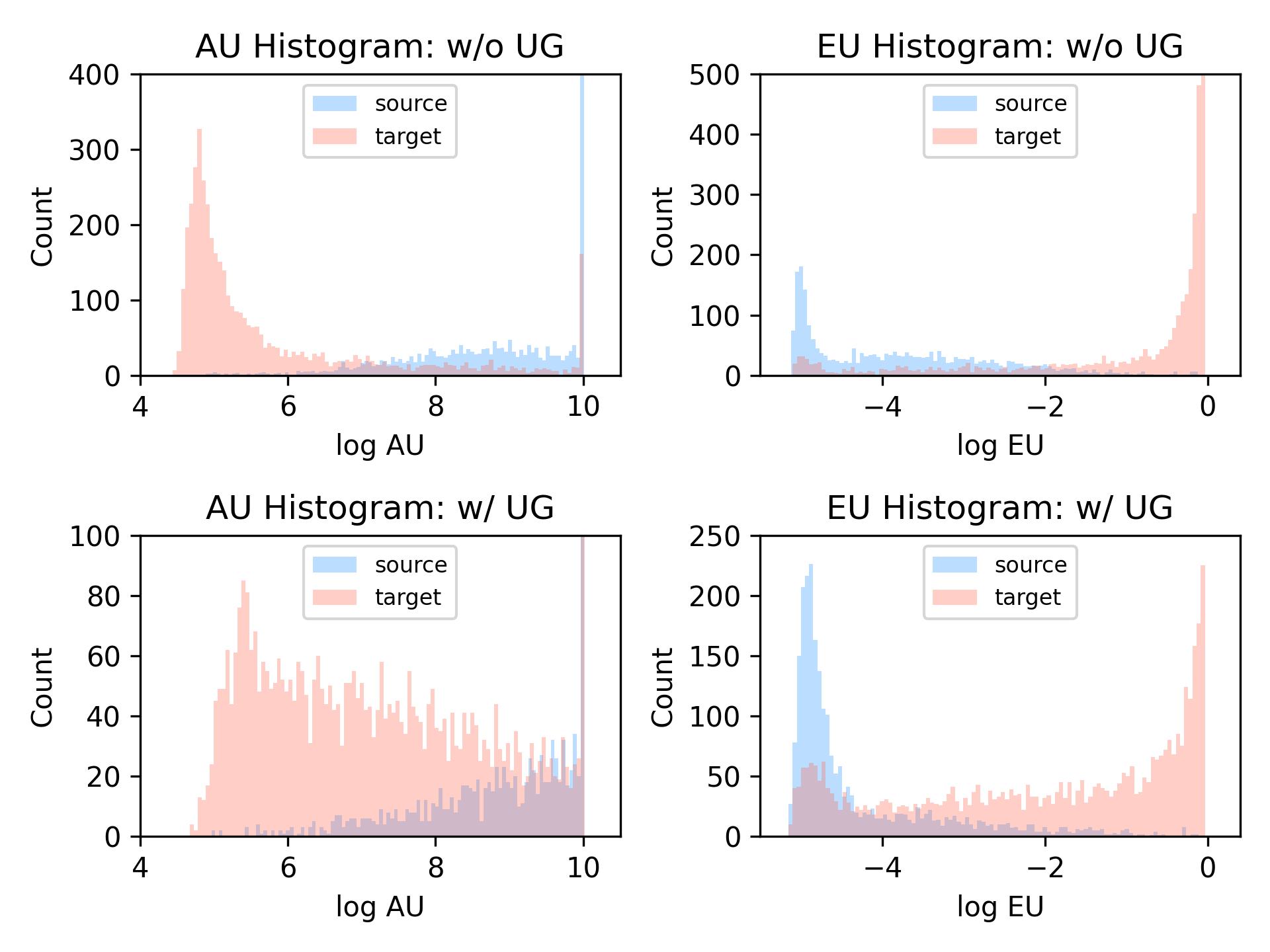}
\caption{Uncertainty histograms of Office-Home source (Art) and target (Clipart) domains. The model is trained on source data only (top row) and with UG (bottom row) without active sampling. Uncertainty is shown in the logarithmic scale for elegancy.}
\label{fig:hist}
\end{figure}

\textbf{Evidential uncertainties across domains.} To investigate whether variance-based epistemic uncertainty is a good measure of targetness and the effect of UG training on domain-level uncertainties, experiments are conducted to obtain the uncertainty histograms demonstrated in \cref{fig:hist}.

The target domain has, in general, a higher epistemic uncertainty when UG is not performed, indicating that epistemic uncertainty can serve as the domain discriminator to some extent. After the model is trained with target uncertainty guidance, epistemic uncertainties of some samples from both domains are reduced, but the difference between the two distributions is still significant.

The aleatoric uncertainty, however, is much smaller in target than in source. When no target label is provided, the model does not have sufficient information to precisely estimate the intrinsic data uncertainty and tends to give relatively low aleatoric uncertainties to the target samples.

\begin{figure}[t]
\centering
\includegraphics[scale=0.32]{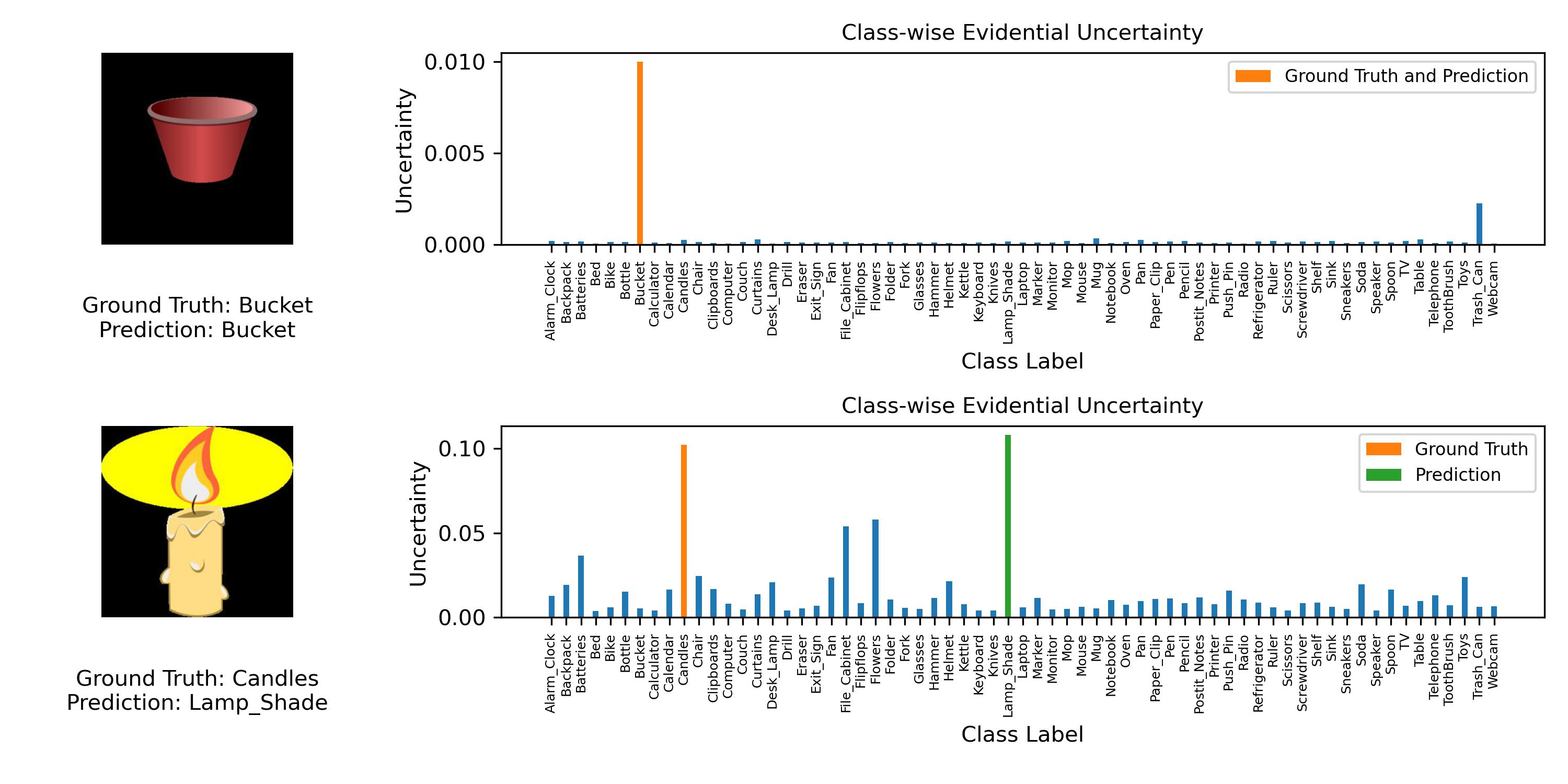}
\caption{
Class-level evidential uncertainties of two example Clipart images from Office-Home A$\rightarrow$C domain transition.}
\label{fig:sample_classU}
\end{figure}

\begin{figure}[t]
\centering
\includegraphics[scale=0.3]{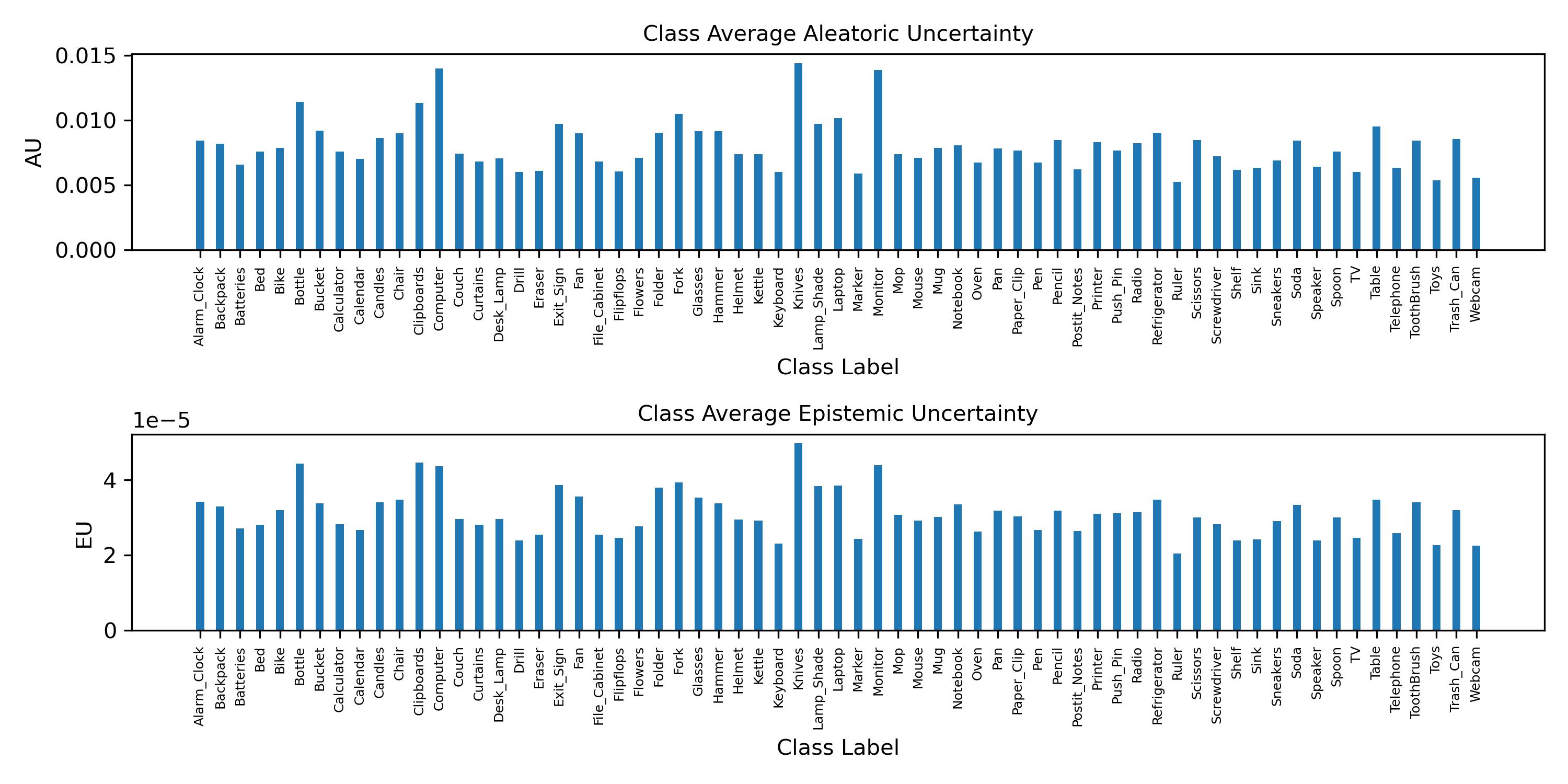}
\caption{Office-Home Clipart domain class-level evidential uncertainties under A$\rightarrow$C domain transition.}
\label{fig:classU}
\end{figure}

\textbf{Class-level Evidential Uncertainties.} Based on \cref{eq:var_classU}-\cref{eq:var_classUepis}, uncertainties can be quantified at the class level for each image in the target domain, showing which classes the model is most uncertain about.
\cref{fig:sample_classU} gives two examples with their class-wise uncertainties.
By aggregating all the domain samples, average class-wise evidential uncertainties can be quantified for the entire domain, as shown in \cref{fig:classU}. This information is useful in applications that require uncertainty estimation at the class level.

\subsubsection{Between-class Correlation Analysis}

The correlation matrix in \cref{eq:corr} can be calculated for each individual image sample $\boldsymbol{x}$. To investigate the correlation between a pair of classes $(i, j)$ in a given dataset, we take the average of $\mathrm{Corr}[\boldsymbol{y}]_{i,j}$ of all the samples from class $i$ and class $j$. This gives the dataset-level class correlation. Samples from classes other than $i$ and $j$ are excluded because most of them do not contain information about the two classes. Note that when class labels are unavailable, predictions can be used as a proxy for ground truth classes.

Here, we assume the labels are available and check the class correlations in the Office-Home Clipart dataset (\cref{tab:rank}). A relatively large negative correlation indicates that the model tends to make more mistakes in distinguishing the two classes. The average correlations are generally not too strong because for many samples, the model is still able to discriminate between the two categories. However, the relative ranking of pair correlations is still informative, as highly similar pairs, such as computer and monitor, can be successfully identified using our approach.

\section{Applications and Limitations}

The EDL-based uncertainty quantification framework is widely applicable in scenarios where uncertainty estimation is crucial. In active learning, the uncertainty metric helps to identify informative samples, which can be annotated to improve model performance. In medical image classification and segmentation, uncertainty indicates to what extent the diagnosis can be trusted. In autonomous driving, uncertainty given by the object detector is pivotal for the vehicle to make driving decisions to ensure safety.

However, the EDL approach is only applicable in the supervised setting, where the ground truth targets are available for model training. For unsupervised learning, EDL is not capable of deriving uncertainties. Extending EDL to unsupervised settings is a possible direction of future research.

\begin{table}[t]
\centering
\resizebox{!}{.04\textheight}{%
\begin{tabular}{c | c c | c}
    \hline \hline
    Rank & Class $i$ & Class $j$ & Correlation \\ \hline
    1 & Computer & Monitor & -0.364 \\ \hline
    2 & Bucket & Trash Can & -0.239 \\ \hline
    3 & Desk Lamp & Lamp Shade & -0.226 \\ \hline
    4 & Knives & Scissors & -0.213 \\ \hline \hline
\end{tabular}
}
\caption{The top 4 most correlated classes in the Office-Home Clipart domain, obtained from A$\rightarrow$C domain transition.} 
\label{tab:rank}
\end{table}

\section{Conclusion}

Inspired by the theory of deep evidential regression, we introduce a variance-based approach to quantify evidential uncertainties in deep learning classification problems.
Compared with the traditional entropy-based approach, our method derives not only sample-level but also class-level uncertainties.
In addition, we show that between-class correlations can be calculated in the EDL setting to identify highly similar class pairs.
We also propose certainty sampling to boost active learning performance under large domain shift.
This variance-based approach will enable the development of more advanced algorithms based on class-level evidential uncertainties and correlation information.

\small
\bibliographystyle{ieee_fullname}
\bibliography{egbib}

\end{document}